\documentclass[lettersize,journal]{IEEEtran}
\usepackage{amsmath,amsfonts}
\usepackage{array}
\usepackage[caption=false,font=normalsize,labelfont=sf,textfont=sf]{subfig}
\usepackage{textcomp}
\usepackage{stfloats}
\usepackage{url}
\usepackage{verbatim}
\usepackage{graphicx}
\usepackage{cite}
\usepackage[ruled]{algorithm2e}
\usepackage{amssymb}

\usepackage[pagebackref,breaklinks,colorlinks]{hyperref}

\usepackage[capitalize]{cleveref}
\crefname{section}{Sec.}{Secs.}
\Crefname{section}{Section}{Sections}
\Crefname{table}{Table}{Tables}
\crefname{table}{Tab.}{Tabs.}

\usepackage{booktabs}
\usepackage{pifont}
\usepackage{multirow}
\usepackage{bm}
\usepackage[table,xcdraw]{xcolor}

\begin{document}

\title{A New Learning Paradigm for Foundation Model-based Remote Sensing Change Detection}

\author{Kaiyu Li, Xiangyong Cao, Deyu Meng
\thanks{Kaiyu Li and Xiangyong Cao are with the School of Computer Science and Technology and Ministry of Education Key Lab For Intelligent Networks and Network Security, Xi’an Jiaotong University, Xi’an 710049, China (email: likyoo.ai@gmail.com, caoxiangyong@xjtu.edu.cn) \textit{(Corresponding author: Xiangyong Cao)}.}
\thanks{Deyu Meng is with the School of Mathematics and Statistics and Ministry of Education Key Lab of Intelligent Network Security, Xi’an Jiaotong University, Xi’an 710049, China (email: dymeng@mail.xjtu.edu.cn).}
}

\markboth{Journal of \LaTeX\ Class Files,~Vol.~14, No.~8, August~2023}%
{Shell \MakeLowercase{\textit{et al.}}: Bare Demo of IEEEtran.cls for IEEE Journals}



\maketitle

\renewcommand{\thefootnote}{\fnsymbol{footnote}}

\begin{abstract}
Change detection (CD) is a critical task to observe and analyze dynamic processes of land cover. Although numerous deep learning-based CD models have performed excellently, their further performance improvements are constrained by the limited knowledge extracted from the given labelled data. On the other hand, the foundation models that emerged recently contain a huge amount of knowledge by scaling up across data modalities and proxy tasks. In this paper, we propose a Bi-Temporal Adapter Network (BAN), which is a universal foundation model-based CD adaptation framework aiming to extract the knowledge of foundation models for CD. The proposed BAN contains three parts, i.e. frozen foundation model (e.g., CLIP), bi-temporal adapter branch (Bi-TAB), and bridging modules between them. Specifically, BAN extracts general features through a frozen foundation model, which are then selected, aligned, and injected into Bi-TAB via the bridging modules. Bi-TAB is designed as a model-agnostic concept to extract task/domain-specific features, which can be either an existing arbitrary CD model or some hand-crafted stacked blocks. Beyond current customized models, BAN is the first extensive attempt to adapt the foundation model to the CD task. Experimental results show the effectiveness of our BAN in improving the performance of existing CD methods (e.g., up to 4.08\% IoU improvement) with only a few additional learnable parameters. More importantly, these successful practices show us the potential of foundation models for remote sensing CD. The code is available at \url{https://github.com/likyoo/BAN} and will be supported in our Open-CD\footnote[2]{An open source CD toolbox: \url{https://github.com/likyoo/open-cd}}.
\end{abstract}

\begin{IEEEkeywords}
Change detection, foundation model, visual tuning, remote sensing image processing, deep learning.
\end{IEEEkeywords}

\section{Introduction}

\IEEEPARstart{T}{he} dynamic properties of the Earth's surface are influenced by a wide range of natural and anthropogenic factors, resulting in constant change. Therefore, analyzing the time-series remote sensing (RS) data is an important topic in Earth vision, and change detection (CD) techniques have become an indispensable tool for this task, contributing to the comprehensive interpretation of surface changes. The basic goal of CD is to detect targets or pixels with semantic changes or specific changes between bi-temporal RS images of the same region. Further, the CD can detect, quantify and analyze changes in land cover and has rich applications in several scenarios, e.g., urban expansion~\cite{chen2020spatial, ji2018fully}, natural disaster assessment~\cite{hansch2022spacenet}, cropland protection \cite{liu2022cnn}, etc.

Traditional CD methods are based on manual feature extraction, making it difficult to achieve fast and accurate detection in complex scenarios \cite{bruzzone2000automatic}. In the last decade, deep learning (DL) technologies have greatly contributed to the development of the field of CD. For example, convolutional neural network (CNN) has achieved practical success in CD, and its superior learning ability and automatic feature extraction capability make it powerful to handle complex scenarios \cite{daudt2018fully, chen2020spatial, fang2021snunet, zhang2020deeply, liu2020building}. Recently, the vision transformer (ViT) \cite{dosovitskiy2020image} has further energized the field of CD \cite{bandara2022transformer, zhang2022swinsunet} since it can handle arbitrary unstructured data and capture global dependencies in the whole image, which opens up more possibilities for CD. Furthermore, some studies have combined CNNs and ViTs in series or parallel form to explore some more powerful CD capabilities \cite{chen2021remote, li2022transunetcd, li2023convtransnet, fu2023slddnet, tang2023wnet}. In summary, these models have achieved practical success to some extent.

However, as the model scale grows, the limited data restricts the performance improvement of the CD models. For instance, the LEVIR-CD dataset \cite{chen2020spatial}, a CD benchmark containing 637 pairs of bi-temporal RS images, on which the DL-based methods can achieve 90\% $F_1$-score by 2020. However, in the following 3 years, the improvement is only $\sim$1\% to 2\%, while the model parameter has grown to hundreds of millions. We consider this to be a performance bottleneck under limited data. To alleviate this problem, some studies have been tried in several aspects, e.g., generating more simulation data using generative models \cite{zheng2023scalable}, constructing more image pairs through data augmentation strategies and single-temporal images \cite{chen2021adversarial, zheng2021change}, etc. However, compared to the millions or even billions of annotated data or image-text pairs in natural images, these generated data are still not able to drive the models to achieve the \textit{emergence} about capabilities and \textit{homogeneity} \cite{bommasani2021opportunities}. Recently, the significance of the foundation model has been increasingly recognized in computer vision. Compared to small models customized for specific tasks and datasets, the foundation models \cite{radford2021learning, cherti2023reproducible, liu2023remoteclip} can accumulate general knowledge through large-scale pre-training, thus reducing the dependence on specific labeled data, which inspires us to introduce the foundation models into the CD task.

The foundation model refers to models that have been trained on large-scale datasets and are capable of capturing a wide range of knowledge and understanding. The foundation model started with natural language processing (NLP) tasks, typically exemplified by OpenAI's GPT series \cite{2023arXiv230308774O}, and has been widely explored in computer vision and vision-language tasks, e.g. CLIP \cite{radford2021learning}, BLIP \cite{li2022blip}, SAM \cite{kirillov2023segment}, etc. These large foundation models allow for cross-domain knowledge transfer and sharing, and reduce the demand for task-specific training data \cite{mai2023opportunities}. In other words, even though in most tasks (e.g., CD) it is difficult to train a task-specific large model from scratch due to the limited data and computational source, we can resort to the existing foundation models and adapt them to specific tasks. Some researchers have tried to tune or adapt foundation models to some downstream tasks in natural images and have achieved some success \cite{chen2022vision, yin20231}. For remote sensing CD, there are two particular issues for its visual tuning: 1) the transfer of the natural image domain to the RS image domain, and 2) the data structure of the input bi-temporal images changes the overall pipeline of foundation models.

\begin{figure}[t]
  \centering
   \includegraphics[width=1.0\linewidth]{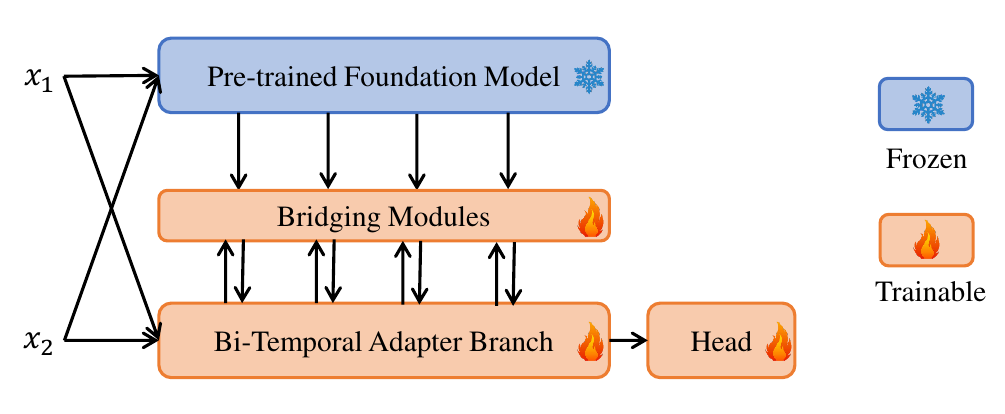}
   \caption{Schematic diagram of BAN. The frozen pre-trained model can be any foundation model (ImageNet-21k pre-trained models \cite{steiner2021train}, CLIP \cite{radford2021learning, cherti2023reproducible}, RemoteCLIP \cite{liu2023remoteclip}, etc.), the Bi-TAB can be any CD model (BiT \cite{chen2021remote}, ChangeFormer \cite{bandara2022transformer}, etc.), and the bridging modules can inject the knowledge extracted from the foundation model into the Bi-TAB.}
   \label{fig:ban}
\end{figure}

To alleviate the two issues, in this paper, we propose a universal framework, i.e., bi-temporal adapter networks (BAN for short), for tuning and adapting foundation models to the CD task. By bridging and combining the foundation model with customized models or modules for CD, BAN can dramatically improve the performance of existing CD models with only a few additional learnable parameters. Specifically, BAN parallel feeds bi-temporal images into a frozen foundation model (i.e., in the siamese-style fashion), which can be ViTs trained on ImageNet-21k \cite{steiner2021train}, the image encoder of CLIP \cite{radford2021learning, cherti2023reproducible}, or RemoteCLIP trained on RS data \cite{liu2023remoteclip}, as well as any other arbitrary pretrained image models. Simultaneously, the bi-temporal images are fed into a bi-temporal adapter branch (Bi-TAB), which can be either an existing CD model e.g. BiT \cite{chen2021remote}, ChangeFormer \cite{bandara2022transformer}, etc., or some hand-crafted stacked blocks. In addition, to alleviate the conflicts in input resolutions \cite{xu2023side} of the foundation model and Bi-TAB, the asymmetric resolution input strategy (ARIS) is proposed. Between the foundation model and the Bi-TAB, we design a series of bridging modules which constantly select, align and inject the general features in the foundation model space into the learnable Bi-TAB at various hierarchies. Through these bridging modules, BAN achieves knowledge transfer from general features to the CD task in the RS image domain. The schematic diagram of BAN is shown in \cref{fig:ban}.

In summary, the contributions of this work are threefold:

\begin{itemize}
\item To reduce the dependence on CD-specific data, we propose a universal framework (i.e., BAN) to adapt the foundation model to the CD task, which efficiently reuses the general knowledge of the foundation model and transfers it to the CD task. Additionally, as a side-tuning framework, BAN allows efficient tuning of parameters, memory and time. To our knowledge, this should be the first universal framework to introduce the foundation models into the CD task. 

\item To better select general features and align them with task/domain-specific features, we propose a bridging module between the foundation model and Bi-TAB. By cross-domain dot-product attention, the bridging module can resample general domain knowledge and then inject it into features of the remote sensing CD domain.

\item Bi-TAB is proposed as a model-agnostic concept, and this branch can be either an existing CD model or some hand-crafted stacked blocks. Benefiting from its plug-and-play property, BAN can be fully equipped by almost any CD model that exists currently. Experimental results on the benchmark CD datasets demonstrate that BAN can bring about 4.08\% improvement on average for the corresponding customized CD models.


\end{itemize}

The rest of this article is organized as follows. In Section II, a brief review of related work is presented. Section III describes the specifics of our proposed framework. Section IV provides a series of experimental results and analysis. Finally, in Section V, we give a conclusion of this paper.

\section{Related work}

\subsection{DL-based CD}

An accurate and reliable CD is essential in the analysis of RS images. Over the past few years, numerous CD approaches have been proposed. In particular, CNN-based methods have shown remarkable success in this task. Daudt et al. \cite{daudt2018fully} proposed three U-Net-based fully convolutional siamese networks: FC-EF, FC-Siam-conc, and FC-Siam-diff, which was one of the earliest studies for DL-based CD. Expanding on the FC-Siam-conc model, Zhang et al. \cite{zhang2020deeply} used the attention mechanism to construct stronger basic blocks, and additionally introduced deep supervision for faster convergence and better performance of the model. Different from the former, STANet \cite{chen2020spatial} uses the metric learning-based approach for CD, which projects the bi-temporal images into a high-dimensional space by a neural network, and then uses the L2-distance metric for the bi-temporal high-dimensional mappings to generate the change mask. In \cite{fang2021snunet}, Fang et al. considered that continuous down-sampling in the backbone of previous models leads to loss of spatial information, resulting in ambiguity in the pixels at the edges of the change targets, and therefore proposed SNUNet, which maintains high-resolution, fine-grained representations through dense skip connections between each node in the encoder and decoder.



The emergence of ViT provides an alternative path for CD tasks. Specifically, ViT splits an image into a series of patches, takes all patches as a sequence, and uses the attention mechanism to capture the global dependencies between patches in the image. Zhang et al. \cite{zhang2022swinsunet} used swin transformer blocks \cite{liu2021swin} to construct an encoder and decoder and designed a pure transformer network with siamese U-shaped structure. ChangeFormer \cite{bandara2022transformer} is a Transformer-based CD model that references the construction of Segformer \cite{xie2021segformer}, a model for semantic segmentation. In the basic module of ChangeFormer, spatial down-sampling of the Query, Key and Value is performed in the self-attention to reduce computation. 


 

Several studies have found that in CD, due to limited data, the pure Transformer model may not reach its full potential without some inductive bias incorporated. Chen et al. \cite{chen2021remote} believed that high-level change features can be represented by a few semantic tokens and proposed BiT. In BiT, the bi-temporal features obtained through ResNet \cite{he2016deep} are represented as several tokens, then, the context is modeled in the token-based space-time using the Transformer encoder, and finally, the tokens with the learned context are fed into the pixel space and refine the original features by the Transformer decoder. Wu et al. \cite{wu2023cstsunet} also used ResNet firstly to extract the bi-temporal features, and then transmitted the bi-temporal features to the designed cross swin-transformer for difference feature extraction. On the other hand, some studies use parallel CNN and Transformer structures. Tang et al. \cite{tang2023wnet} proposed WNet, which uses a siamese CNN and a siamese Transformer as encoders and fuses all features in the decoder. Similarly, Feng et al. \cite{feng2022icif} proposed ICIF-Net, which constructs a dual-branch structure of CNN and Transformer to capture multiscale local and global features, respectively, and uses cross-attention to fuse them. Fu et al. \cite{fu2023slddnet} argued that there is no interaction between dual parallel branches in the feature extraction processes of ICIF-Net, so they designed a semantic information aggregation module to dynamically implement the interaction between the CNN and Transformer branches in their proposed SLDDNet. 



In this work, beyond the above customized CD models, we propose a universal CD framework (i.e. BAN) based on the foundation model. In general, all the above-mentioned CD models can be used as the Bi-TAB in BAN, and BAN can help improve the performance of these existing CD models.

\subsection{Foundation Model and Visual Tuning}

Through various pre-training techniques, the foundation model can be scaled up in data modalities, proxy tasks, etc., thus accumulating knowledge \cite{yu2023visual}, which has gradually become a consensus among researchers. With the success of ViT, it swept the field of computer vision and became the priority choice for the vision foundation models. Steiner et al. \cite{steiner2021train} conducted the first systematic study of the interplay between data, augmentation, and regularization when pre-training ViT. CLIP is a pre-training method based on contrastive language-image
learning, it is trained to maximize the similarity between the vision and language encodings of matching image-text pairs while minimizing it for non-matching pairs. With strong generalization ability and extensibility, CLIP can handle multiple types of image and text data without retraining for specific tasks \cite{xu2023side}. Inspired by the CLIP, Liu et al. \cite{liu2023remoteclip} proposed RemoteCLIP, the first visual-language foundation model for RS. Although RemoteCLIP has achieved some improvement in some RS tasks, it is still limited by the volume of data, i.e., it contains only 828k image-text pairs. For this, Zhang et al. \cite{zhang2023rs5m} proposed the RS5M dataset containing 5 million RS images with English descriptions by filtering the image-text pair dataset and generating captions for RS images \cite{laion_coco}. 





How to adapt the above-mentioned foundation models to downstream specific tasks is an important topic currently. Typical transfer learning methods such as fully fine-tuning the whole model or only fine-tuning the task head, lead to high training costs or insufficient reuse, and one trade-off is the parameter-efficient transfer learning (PETL) method \cite{yu2023visual}, which selects a subset of pre-trained parameters or/and introduces a limited number of trainable parameters into the foundation model while freezing the majority of the original parameters \cite{sung2022lst}. Typical PETL methods include prompt tuning, which concatenates some learnable embeddings with inputs, and adapter, which inserts some learnable components into the pre-trained model. Hu et al. \cite{hu2021lora} proposed LoRA, which initializes a low-rank adaptation matrix and inserts it into self-attention in the residual-connected form. Due to the simplicity and effectiveness of LoRA, numerous related studies and practices have emerged \cite{yin20231}. However, parameter efficiency does not mean memory efficiency, and the foundation model tuning remains difficult. To address this issue, Sun et al. \cite{sung2022lst} proposed the ladder side-tuning (LST) method, which freezes the foundation model and builds a side network for training \cite{zhang2020side, xu2023side, lin2023hierarchical}. In this paper, our proposed BAN can be seen as the practice and extension of side-tuning for the CD task.




Most relevant to this paper is a concurrent work, SAM-CD \cite{ding2023adapting}, which uses frozen Fast-SAM \cite{zhao2023fast} as the encoder and fine-tunes the feature pyramid network and prediction head for CD. However, compared to BAN, SAM-CD is more of a customized Fast-SAM-based model within a traditional transfer learning paradigm. Therefore, we claim that BAN should be the first universal framework to adapt the foundation model to the CD task.

\begin{figure*}[t]
  \centering
   \includegraphics[width=0.95\linewidth]{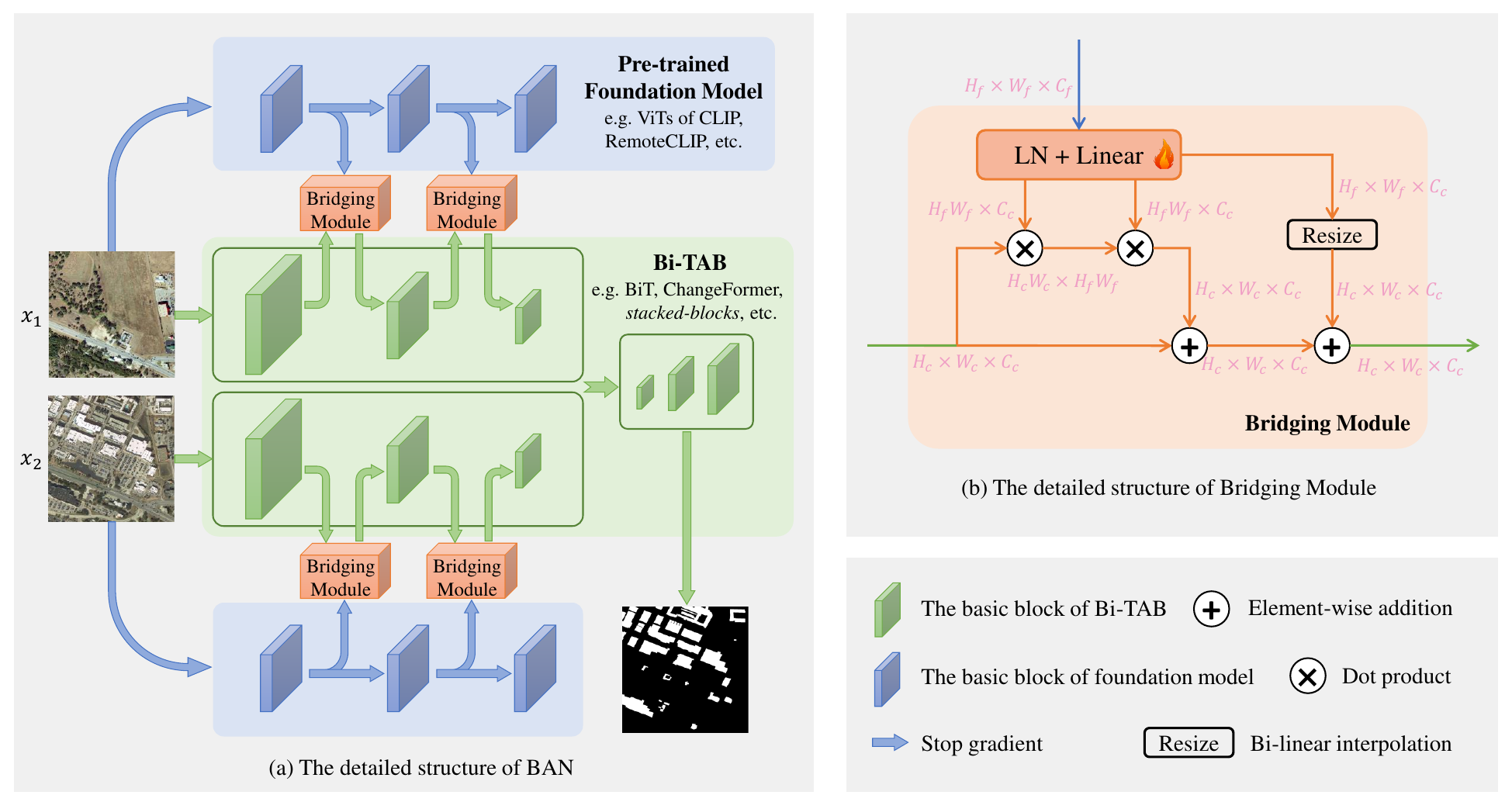}
   \caption{\textbf{The detailed illustration of BAN.} The foundation model (blue area) accumulates general knowledge through pre-training techniques, and the bridging modules (orange area) select, align and inject this knowledge into the Bi-TAB. The Bi-TAB (green area) is a model-agnostic concept, which can be an arbitrary customized CD model or even some hand-crafted stacked blocks. These three major components are detailed in Section \ref{section:FM}, \ref{section:Bi-TAB} and \ref{section:BM}, respectively.}
   \label{fig:BAN_detail}
\end{figure*}

\section{Method}

As illustrated in \cref{fig:ban} and \cref{fig:BAN_detail}, BAN can be divided into three parts: frozen foundation model, Bi-TAB and bridging modules between them. For the foundation model, it is generally ViTs with various scales. Next, we will review the structure of ViT. Then, we will introduce Bi-TAB and explain why Bi-TAB can be an existing arbitrary CD model. Finally, we will present the details of the bridging module.


\subsection{Review of ViT}
\label{section:FM}

For an image $x \in \mathbb{R}^{H \times W \times C}$, ViT first divide it into a series of non-overlapping $P \times P$ patches, where $P$ is typically 14, 16, and 32. Each patch is encoded into a D-dimensional vector using a convolutional layer, and the output $x_{p}\in \mathbb{R}^{N \times D}$ is termed patch embedding, where $ N = HW/P^2$, and $D$ denotes the constant dimension of ViT. Then, a learnable class token is concatenated into $x_p$ as the whole image representation. Due to the insensitivity to sequence order, a learnable position encoding is subsequently added to the patch embedding.

The basic block of ViT consists of alternating multi-head self-attention (MSA) and feed-forward network (FFN), with layer normalisation (LN) and residual connection applied in each layer. Formally, a basic block can be expressed as:

\begin{equation}
\begin{aligned}
  & x_l^{\prime} = MSA(LN(x_{l-1})) + x_{l-1}, \\
  & x_l = FFN(LN(x_l^{\prime})) + x_l^{\prime}, \hspace{2em} l=1,2,...,L
  \label{eq:1}
\end{aligned}
\end{equation}
where $L$ denotes the number of basic blocks in ViT, $x_l$ denotes the output of the $l$-th block, the FFN consists of two linear layers with a GELU non-linearity, and the MSA can be formulated as:

\begin{equation}
\begin{aligned}
  MSA(Q,K,V) &= Concat(head_1,...,head_h)W^O, \\
  head_i &= SA(QW_i^Q, KW_i^K, VW_i^V), \\
  SA(Q,K,V) &= softmax(\frac{QK^{\mathsf{T}}}{\sqrt{D}})V,
  \label{eq:2}
\end{aligned}
\end{equation}
where $Q$, $K$ and $V$ denote Query, Key and Value, respectively, and $W_i^Q$, $W_i^K$, $W_i^V$ and $W^O$ are projection parameters.

In BAN, since two images (bi-temporal images) need to be received as inputs, we extract bi-temporal features simultaneously using a parallel frozen foundation model with shared parameters, i.e. a siamese foundation model. High-resolution images are important for accurate CD, however, most foundation models are trained on low-resolution images (e.g., $224 \times 224$ or $336 \times 336$). On the other hand, the position embedding of ViT is highly correlated with the image resolution, and if there is an image resolution difference in inference, a new approximate position encoding needs to be recomputed by interpolation. In addition, self-attention does not have the translation equivariance that exists in convolution and thus is more sensitive to resolution. Therefore, ARIS is proposed in BAN. Specifically, for the foundation model, the bi-temporal images are resized to match their pre-training resolution, while for Bi-TAB, the resolution is kept as $256 \times 256$ or $512 \times 512$ according to the dataset.

Different from task-specific models that are trained on specific datasets, foundation models are trained in a self-supervised or semi-supervised manner on large-scale data and can adapt their knowledge and capabilities to several downstream tasks. Under similar ViT architectures, a crucial difference between various foundation models is the training data and training strategy, in this paper, we mainly use CLIP's image encoder (ViT-L/14) \cite{radford2021learning, cherti2023reproducible} as the foundation model part in BAN. In the experimental part, we compare the foundation models under different weights and architectures, and the details can be found in Section \ref{section:ablation_studies_dfm}.

\subsection{Bi-TAB}
\label{section:Bi-TAB}

Since the foundation model part is frozen, it can only extract general information from the image. However, this information is incomplete and Bi-TAB is thus needed for domain-specific and task-specific feature extraction. Bi-TAB is not designated as an explicit and concrete network, but rather as a model-agnostic concept. Therefore, Bi-TAB can be either a customized CD model, including existing CNN-based and Transformer-based models, or several hand-crafted stacked blocks. In our work, we recommend the former since it allows BAN to better benefit from evolving customized CD models.

\begin{figure}[t]
  \centering
   \includegraphics[height=0.93\linewidth]{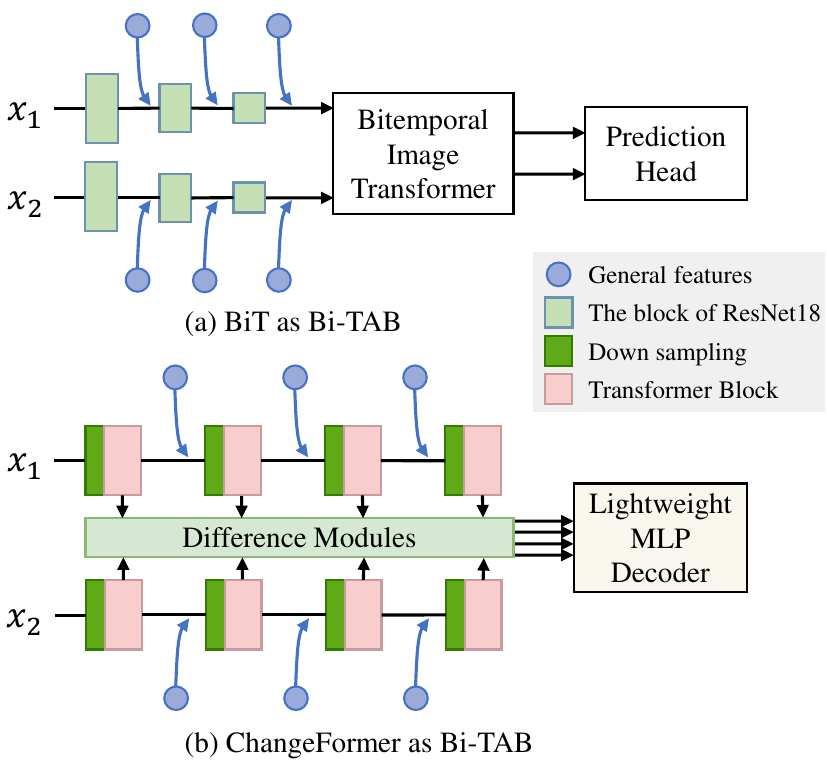}
   \caption{Illustration of the Bi-TAB perspective in BAN, with BiT \cite{chen2021remote} and ChangeFormer \cite{bandara2022transformer} as examples. For better presentation, the color renderings follow their original literature \cite{chen2021remote, bandara2022transformer}.}
   \label{fig:bi-tab}
\end{figure}

A modern customized CD model usually contains a backbone (encoder) and a prediction head (decoder), and its backbone is always a Siamese network. Thus, a customized CD model can simply be represented as

\begin{equation}
\begin{aligned}
  \hat{Y} = Head(Backbone(x_1), Backbone(x_2)),
  \label{eq:3}
\end{aligned}
\end{equation}
where $x_1$ and $x_2$ denote the bi-temporal images, $\hat{Y}$ denotes the change mask, and the $Backbone$ and the $Head$ consists of a sequence of basic blocks as shown in \cref{fig:BAN_detail}(a). In the perspective of Bi-TAB, as shown in \cref{fig:bi-tab}, except for the original bi-temporal images, an additional general feature bank is provided by the foundation model, from which Bi-TAB can arbitrarily take out features. Since both are siamese structures, information from the foundation model can be easily injected into the Bi-TAB's backbone of the corresponding temporal phase, which is why almost any CD model can be chosen as the Bi-TAB. With the support of the foundation model, Bi-TAB can easily obtain general information about the image, which is typically obtained from a large amount of feeding data. In this way, the requirement of the model for data is reduced.

As presented in \cref{fig:BAN_detail}(a) and \cref{fig:bi-tab}, modern customized CD models can be easily embedded into BAN. Specifically in this paper, we mainly use two typical CD customized models as Bi-TAB, CNN-Transformer-based BiT \cite{chen2021remote} and Transformer-based ChangeFormer \cite{bandara2022transformer} (denoted by ``CF''). Among them, in the original CF, the backbone network is a customized mix transformer (MiT) \cite{xie2021segformer}, which can help disable the model scalability. Therefore, in this paper, we use the scalable MiT-b0 to MiT-b5 settings in SegFormer \cite{xie2021segformer}. Furthermore, to validate the performance of BAN in the semantic change detection (SCD) task, we remodel CF into CF-SCD by adding two additional semantic segmentation prediction heads.

\subsection{Bridging Module}
\label{section:BM}

Next, a natural question is how to inject the general features from the foundation model into Bi-TAB, and we found that there are two critical issues: 1) not all the general knowledge is needed, so we should consider how to filter out the useless information from the general knowledge and select the valuable information. 2) With ARIS and patch embedding, the features in the foundation model have extremely low resolution (being downsampled by at least $14\times$), so it is an important issue to align the features from both sources. Due to the above two issues, we propose the bridging module, which is shown in \cref{fig:BAN_detail}(b).

There are two types of features in our framework, i.e. the general feature $x^{fm}\in \mathbb{R}^{H_fW_f \times C_f}$ of the foundation model and the task-specific feature $x^{cm}\in \mathbb{R}^{H_cW_c \times C_c}$ from the Bi-TAB. In the bridging module, to avoid the inconsistency of the distribution between $x^{fm}$ and $x^{cm}$, we first use an LN layer to normalize $x^{fm}$, and then a linear layer is used to project $x^{fm}$ to $\tilde{x}^{fm} \in \mathbb{R}^{H_fW_f \times C_c}$:

\begin{equation}
\begin{aligned}
  \tilde{x}^{fm} = Linear(LN(x^{fm})).
  \label{eq:BM_1}
\end{aligned}
\end{equation}

To effectively discriminate valuable characteristics in general knowledge, we calculate an affinity matrix $A\in \mathbb{R}^{H_cW_c \times H_fW_f}$ between $\tilde{x}^{fm}$ and $x^{cm}$ using the cosine metric and obtain the attention weights $\tilde{A}$ by dividing $A$ by $\sqrt{C_c}$ and applying a softmax function. Then, the attention weight $\tilde{A}$ is dot-producted with $\tilde{x}^{fm}$ to obtain the filtered feature ${x}^{cf}$:

\begin{equation}
\begin{aligned}
  {x}^{cf} = softmax(\frac{(x^{cm})(\tilde{x}^{fm})^{\mathsf{T}}}{\sqrt{C_c}})\tilde{x}^{fm}.
  \label{eq:BM_2}
\end{aligned}
\end{equation}

It is worth noting that the foundation model feature $x^{fm}$ is resampled to ${x}^{cf} \in \mathbb{R}^{H_c W_c \times C_c}$ in this process, which also solves the scale misalignment problem. Finally, the residual connections from $x^{cm}$ and $\tilde{x}^{fm}$ are adopted to get the ultimate output $x^{bm}$ of the bridging module, which is also the input of the next basic block of Bi-TAB:

\begin{equation}
\begin{aligned}
  x^{bm} = {x}^{cf} + Resize(\tilde{x}^{fm}) + {x}^{cm},
  \label{eq:BM_3}
\end{aligned}
\end{equation}
where the $Resize()$ indicates upsampling $\tilde{x}^{fm}$ to $H_c \times W_c \times C_c$ using bi-linear interpolation.

\begin{algorithm} 
\caption{Forward Propagation of BAN } 
// $FM$: Foundation Model \;
// $CM$: Change Detection Model (i.e. Bi-TAB) \;
// $BM$: Bridging Modules \;
// For brevity, here it is assumed that there are $J$ blocks \\
// in both $FM$, $CM$ and $BM$, using $j$ for indexing \;
\KwIn{$X = \{(x_1, x_2)\}$ (a pair of bi-temporal images)} 
\KwOut{$\hat{Y}$ (a prediction change mask)} 
Initialize $FM$ $\leftarrow$ pre-trained weights\;
\For{$i$ in $\{1,2\}$} 
{
    $x_i^{fm} = FM_{patch\_embedding}(x_i)$\;
    $x_i^{fm} = FM_{position\_embedding}(x_i^{fm})$\;
    $x_i^{cm} = CM_{pre}(x_i)$\;
    \For{$j=1$ to $J$}
    {
    $x_i^{fm} = FM_j(x_i^{fm})$ \;
    $x_i^{cm} = CM_j(x_i^{cm})$ \;
    $x_i^{cm} = BM_j(x_i^{fm}, x_i^{cm})$ \;
    }
} 
// obtain change mask by the prediction head \\
$\hat{Y} = CM_{head}(x_1^{cm}, x_2^{cm})$
\end{algorithm}

\subsection{Loss Function}
The loss function of BAN can directly follow the plain loss function of Bi-TAB. Specifically, the pixel-wise cross-entropy loss for BiT and ChangeFormer is adopted in this paper, which is formally defined as:

\begin{equation}
\begin{aligned}
  L_{ce} = - \frac{1}{H\times W} \sum_{h=1,w=1}^{H,W} Y(h,w)log(\hat{Y}(h,w)) 
  \label{eq:loss}
\end{aligned}
\end{equation}
where $H$ and $W$ denote the height and width of the output, and $Y(h,w)$ and $\hat{Y}(h,w)$ denote the label and prediction for the pixel at position $(h,w)$. For SCD with auxiliary tasks, $L_{ce}$ is also applied to the semantic segmentation sub-task.

\section{Experiments}

\begin{figure*}[t]
  \centering
   \includegraphics[width=0.87\linewidth]{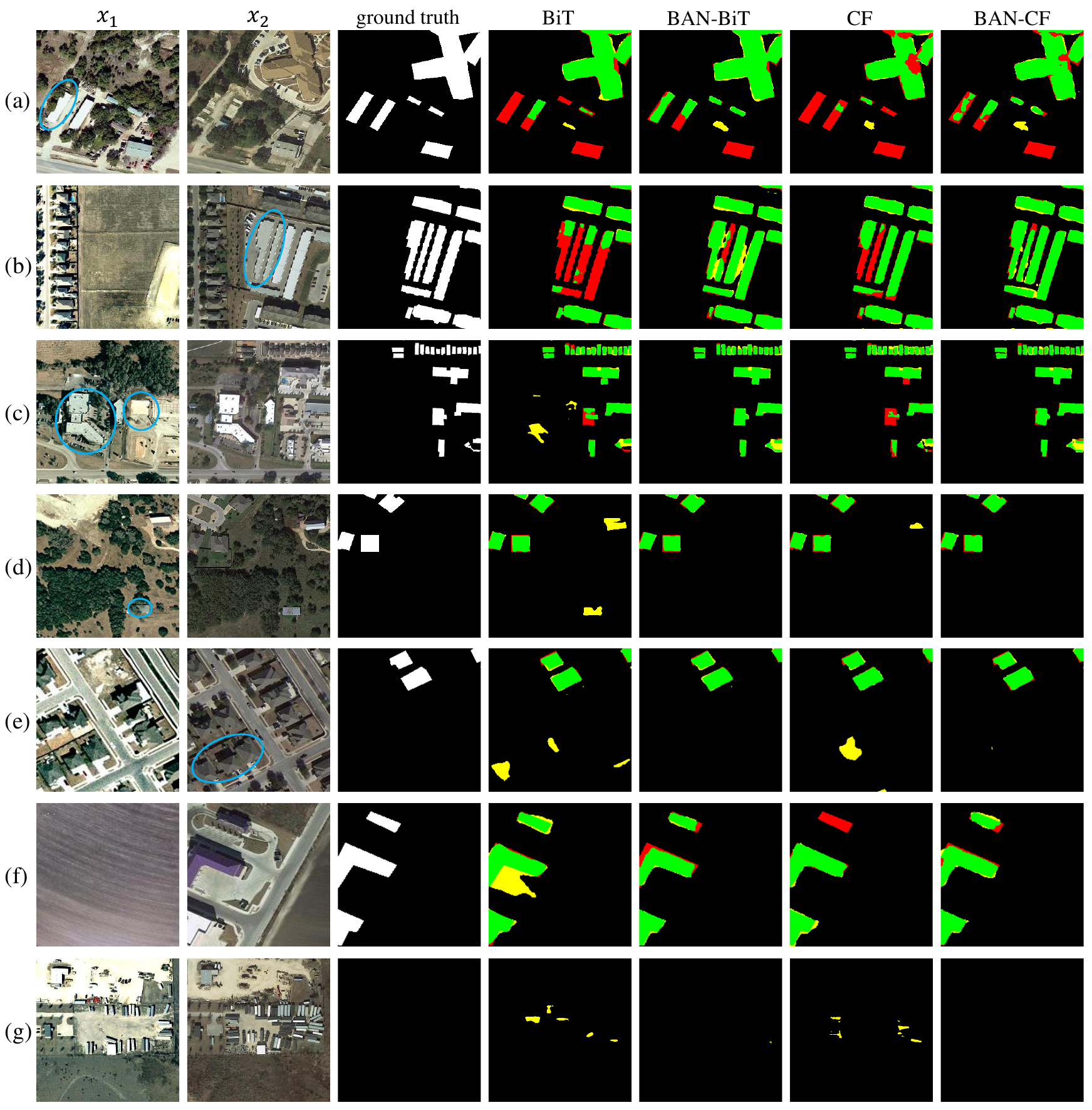}
   \caption{Visualization results of different methods on the LEVIR-CD testing set. The rendered colors represent {\color{green}{true positives (TP)}}, {\color{yellow}false positives (FP)}, {\color{red}false negatives (FN)} and {\color{black}true negatives (FP)}.}
   \label{fig:cd_compare}
\end{figure*}

\subsection{Datasets}

To sufficiently validate the performance of BAN, we conducted extensive experiments on the following five CD datasets consisting of RGB images. It is noted that BAN and all compared methods are trained and evaluated on these datasets without using any extra data.

\begin{itemize}
\item LEVIR-CD dataset \cite{chen2020spatial} comprises a collection of 637 pairs of bi-temporal RS images. These images were obtained from Google Earth and are accompanied by over 31,333 annotated instances of changes. Each image in pairs is $1024 \times 1024$ with a spatial resolution of 0.5 m/pixels. To ensure standardized evaluation, the dataset is divided into three subsets: training, validation, and testing, containing 445, 64 and 128 image pairs.
  
\item S2Looking \cite{shen2021s2looking} is a building CD dataset that contains large-scale side-looking satellite images captured at varying angles from the nadir. It consists of 5,000 co-registered bi-temporal image pairs of rural areas around the world, with the size of $1024 \times 1024$ and spatial resolution of 0.5$\sim$0.8 m/pixel, and in total 65,920 annotated change instances. It provides two annotated maps for each pair of images, indicating new and demolished building areas, which we use to generate a change map.

\item WHU-CD dataset \cite{ji2018fully} contains a pair of images taken in the same area from 2012 and 2016, each with a size of 32,507 × 15,354 pixels. The original images are first cropped into several non-overlapping image patches of size $256 \times 256$. Following \cite{bandara2022revisiting}, these patches are divided into 5947/743/744 for training, validation, and testing, and part of the annotated training data are selected for the semi-supervised case.

\item BANDON dataset \cite{pang2023detecting} contains a large number of off-nadir aerial images with a spatial resolution of 0.6 m/pixels, collected mainly from Google Earth, Microsoft Virtual Earth, and ArcGIS, and the data are divided into 1689/202/392 for training, validation, and testing, with a size of $2048 \times 2048$. To better evaluate the generalization ability of models, the testing set is divided into an in-domain set and an out-domain set (collected from different cities) containing 207 and 185 image pairs. With multiple types of annotations, the BANDON dataset can also be used for the SCD task.

\item The Landsat-SCD dataset \cite{yuan2022transformer} consists of 8,468 image pairs collected between 1990 and 2020, each with a size of 416 × 416 and a spatial resolution of 30 m/pixel. This data set is annotated with a ``no-change'' category and ten types of semantic changes, which contain four classes of land cover (farmland, desert, building, and water). Since there is no official division, we divided the dataset into training, validation and testing sets in the ratio of 3:1:1.

\end{itemize}

\subsection{Implementation details}

We implement our BAN using our Open-CD, which is a PyTorch-based CD toolkit. During training, we use the cross-entropy loss and AdamW optimizer. The learning rate is set to 0.0001, and $10\times$ in Bi-TAB's prediction head, for all models and datasets. The learning rate is decayed using the $poly$ schedule. Following \cite{fang2022changer}, we use random crop (crop size is set to $512\times 512$ for LEVIR-CD, S2Looking and BANDOM datasets; no extra crop for WHU-CD and Landsat-SCD datasets), flip and photometric distortion for data augmentation. Following some common settings, we train 40k, 80k and 40k iterations on the LEVIR-CD, S2Looking and BANDOM datasets, and 100 and 50 epochs on the WHU-CD and Landsat-SCD datasets. All experiments are performed on NVIDIA GeForce RTX 4090 and NVIDIA RTX A6000 GPUs and the batch size is set to 8. If not specified, CLIP's ViT-L/14 is used as the foundation model part of BAN and ARIS is enabled by default.

\begin{figure*}[t]
  \centering
   \includegraphics[width=0.9\linewidth]{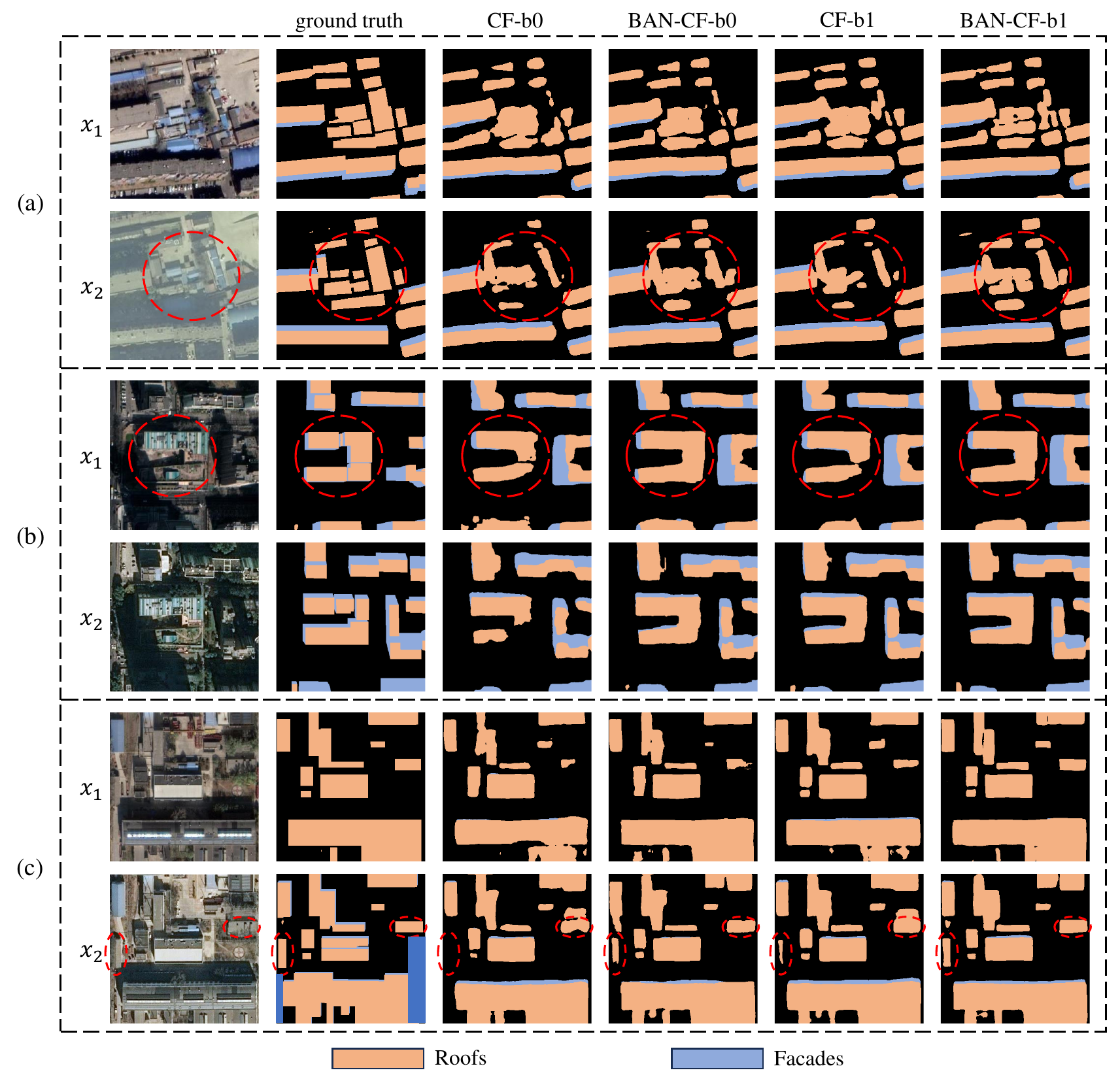}
   \caption{Visualization results (semantic segmentation auxiliary task) of different methods on the BANDON dataset.}
   \label{fig:cd_compare_bandon}
\end{figure*}

\subsection{Evaluation metrics}

For BCD tasks, we use the Intersection Over Union ($IoU^c$), $F_1^c$ score, and Overall Accuracy ($OA$) as evaluation metrics, which are calculated as follows:


\begin{equation}
\label{IoU_c}
IoU^c = \frac{TP}{TP+FP+FN},
\end{equation}

\begin{equation}
\label{F1-Score}
F_1^c = \frac{2TP}{2TP+FP+FN},
\end{equation}




\begin{equation}
\label{OA}
OA = \frac{TP+TN}{FN +FP +TP +TN},
\end{equation}
where TP, TN, FP, and FN indicate true positive, true negative, false positive, and false negative, which are calculated on the change category to avoid the class imbalance problem. $Precision^c$ and $Recall^c$ are also calculated, indicating the precision and recall on the change category.

For the SCD task, we use $mIoU$ and adopt the separated coefficient of kappa ($Sek$) \cite{yang2021asymmetric} as evaluation metrics, which are calculated as follows:

\begin{equation}
\label{mIoU}
mIoU = \frac{IoU^u+IoU^c}{2},
\end{equation}


\begin{equation}
\label{SeK}
SeK = e^{IoU^c-1} \cdot Kappa,
\end{equation}

\begin{equation}
\label{Kappa}
Kappa = \frac{p_o - p_e}{1 - p_e},
\end{equation}
where $IoU^u$ is calculated on the non-change category, corresponding to $IoU^c$. $Kappa$ is calculated for bi-temporal semantic segmentation, $p_o$ denotes the value of the number of correctly predicted samples divided by the total number of samples, and $p_e$ denotes the sum of the products of the number of samples and the number of predictions for all categories divided by the square of the total number of samples. Then, a weighted $Score$ of $Sek$ and $mIoU$ is calculated with weights of 0.7 and 0.3.



\subsection{Comparison and Analysis}
\subsubsection{\textbf{Binary Change Detection (BCD)}}
\label{section:bcd_exp}

\begin{table}
  \caption{Comparisons of BAN with other BCD methods on learnable parameters ($Param$ (M)), $Precision^c$($P^c$) (\%), $Recall^c$($R^c$) (\%), $F_1^c$ (\%) and $IoU^c$ (\%) on LEVIR-CD dataset. The symbol ``*'' means our re-implemented results.}
  \label{table_levircd}
  \centering
  \scalebox{0.9}{
  \begin{tabular}{@{}l|c|c|c|c|c@{}}
    \toprule[1pt]
    Method & $Param$ & $P^c$ & $R^c$ & $F_1^c$ & $IoU^c$\\
    \midrule
    FC-EF \cite{daudt2018fully} & 1.35 & 86.91 & 80.17 & 83.4 & 71.53 \\
    FC-Siam-Diff \cite{daudt2018fully} & 1.35 & 89.53 & 83.31 & 86.31 & 75.92 \\
    FC-Siam-Conc \cite{daudt2018fully} & 1.54 & 91.99 & 76.77 & 83.69 & 71.96 \\
    DTCDSCN \cite{liu2020building} & 41.07 & 88.53 & 86.83 & 87.67 & 78.05 \\
    STANet \cite{chen2020spatial} & 16.93 & 83.81 & \textbf{91.00} & 87.26 & 77.40 \\
    IFNet \cite{zhang2020deeply} & 50.71 & 94.02 & 82.93 & 88.13 & 78.77 \\
    SNUNet \cite{fang2021snunet} & 12.03 & 89.18 & 87.17 & 88.16 & 78.83 \\
    ConvTransNet \cite{chen2021remote} & 7.13 & 91.57 & 90.26 & 90.91 & N/A \\
    WNet \cite{tang2023wnet} & 43.07 & 91.16 & 90.18 & 90.67 & 82.93\\
    CSTSUNet \cite{wu2023cstsunet} & 42.17 & 92.51 & 89.16 & 90.81 & 83.16 \\
    \hline
    BiT* \cite{chen2021remote} & 2.99 & 91.97 & 88.62 & 90.26 & 82.25 \\
    \rowcolor{gray!30}
    BAN-BiT & 3.80 & \textbf{92.83} & 90.89 & \textbf{91.85} & \textbf{84.93} {\color{green!50!black}$\uparrow$\scriptsize{2.68}}\\
    CF \cite{bandara2022transformer} & 41.02 & 92.05 & 88.80 & 90.40 & 82.48 \\
    \rowcolor{gray!30}
    BAN-CF-b0 & 4.34 & \textbf{93.47} & 90.30 & \textbf{91.86} & \textbf{84.94} {\color{green!50!black}$\uparrow$\scriptsize{2.46}} \\
    \hline
    \textit{stacked-blocks} & 1.44 & 91.72 & 87.83 & 89.73 & 81.38 \\
    \rowcolor{gray!30}
    BAN\textit{-stacked-blocks} & 2.14 & 92.72 & 90.64 & 91.66 & 84.61 {\color{green!50!black}$\uparrow$\scriptsize{3.23}} \\
    \bottomrule[1pt]
  \end{tabular}}
\end{table}

\begin{table}
  \caption{Comparisons of BAN with other BCD methods on learnable parameters ($Param$ (M)), $Precision^c$($P^c$) (\%), $Recall^c$($R^c$) (\%), $F_1^c$ (\%) and $IoU^c$ (\%) on S2Looking dataset. The symbol ``*'' means our re-implemented results.}
  \label{table_s2looking}
  \centering
  \scalebox{0.9}{
  \begin{tabular}{@{}l|c|c|c|c|c@{}}
    \toprule[1pt]
    Method & $Param$ & $P^c$ & $R^c$ & $F_1^c$ & $IoU^c$\\
    \midrule
    FC-EF \cite{daudt2018fully} & 1.35 & \textbf{81.36} & 8.95 & 7.65 & N/A \\
    FC-Siam-Diff \cite{daudt2018fully} & 1.35 & 68.27 & 18.52 & 13.54 & N/A \\
    FC-Siam-Conc \cite{daudt2018fully} & 1.54 & \textbf{83.29} & 15.76 & 13.19 & N/A \\
    DTCDSCN \cite{liu2020building} & 41.07 & 68.58 & 49.16 & 57.27 & N/A \\
    STANet \cite{chen2020spatial} & 16.93 & 38.75 & 56.49 & 57.27 & N/A \\
    IFNet \cite{zhang2020deeply} & 50.71 & 66.46 & \textbf{61.95} & 64.13 & N/A \\
    SNUNet \cite{fang2021snunet} & 12.03 & 71.94 & 56.34 & 63.19 & 46.19 \\
    HCGMNet \cite{han2023hcgmnet} & 47.32 & 70.51 & 56.87 & 62.96 & 45.94 \\
    CGNet \cite{han2023change} & 33.99 & 70.18 & 59.38 & 64.33 & 47.41 \\
    \hline
    BiT* \cite{chen2021remote} & 2.99 & 74.80 & 55.56 & 63.76 & 46.80 \\
    \rowcolor{gray!30}
    BAN-BiT & 3.80 & 75.06 & 58.00 & \textbf{65.44} & \textbf{48.63} {\color{green!50!black}$\uparrow$\scriptsize{1.83}} \\
    CF* \cite{bandara2022transformer} & 41.02 & 77.68 & 55.25 & 64.57 & 47.68 \\
    \rowcolor{gray!30}
    BAN-CF-b1 & 15.88 & 74.63 & \textbf{60.30} & \textbf{66.70} & \textbf{50.04} {\color{green!50!black}$\uparrow$\scriptsize{2.36}} \\

    \bottomrule[1pt]
  \end{tabular}}
\end{table}

\begin{table*}
  \caption{Comparisons of BAN with other CD methods on learnable parameters ($Param$ (M)), FLOPs (G), $Precision^c$($P^c$) (\%), $Recall^c$($R^c$) (\%), $F_1^c$ (\%) and $IoU^c$ (\%) on BANDON dataset. The FLOPs are computed  at the resolution of $512\times 512$. The symbol ``*'' means our re-implemented results and the symbol ``\dag'' means that more auxiliary tasks (labels) are used.}
  \label{table_bandon}
  \centering
  \scalebox{0.95}{
  \begin{tabular}{@{}l|c|c|c|cccc|cccc@{}}
    \toprule[1pt]
    \multirow{2}{*}{Method} & \multirow{2}{*}{Backbone} & \multirow{2}{*}{$Param$} & \multirow{2}{*}{FLOPs} & \multicolumn{4}{c|}{In-domain Test} & \multicolumn{4}{c}{Out-domain Test}\\
    & & & & $P^c$ & $R^c$ & $F_1^c$ & $IoU^c$ & $P^c$ & $R^c$ & $F_1^c$ & $IoU^c$\\
    \midrule
    FC-EF \cite{daudt2018fully} & ResNet-50 & 30.75 & 285.50 & 51.80 & 34.83 & 41.65 & 23.30 & 40.76 & 30.76 & 35.04 & 21.24 \\
    FC-Siam-Diff \cite{daudt2018fully} & ResNet-50 & 30.75 & 386.49 & 65.11 & 67.39 & 66.23 & 49.51 & 60.33 & 60.06 & 60.20 & 43.06 \\
    FC-Siam-Conc \cite{daudt2018fully} & ResNet-50 & 41.90 & 434.55 & 65.49 & 66.17 & 65.83 & 49.06 & 60.81 & 58.65 & 59.71 & 42.56 \\
    DTCDSCN \cite{liu2020building} & SE-ResNet-34 & 41.07 & 60.87 & 69.23 & 57.90 & 63.06 & 46.05 & 54.22 & 56.89 & 55.53 & 38.43 \\
    STANet \cite{chen2020spatial} & ResNet-50 & 25.96 & 525.43 & 66.24 & 69.42 & 67.79 & 51.28 & 60.35 & 61.92 & 61.12 & 44.01 \\
    Change-Mix \cite{zheng2021change} & ResNet-50 & - & - & 69.29 & 70.80 & 70.04 & 53.89 & 65.60 & 63.26 & 64.41 & 47.50 \\
    MTGCD-Net\dag \cite{pang2023detecting} & ResNet-50 & 91.85 & 503.53 & 73.92 & 76.55 & 75.21 & 60.27 & 71.52 & 67.30 & 69.35 & 53.08 \\
    \hline
    BiT* \cite{chen2021remote} & ResNet-18 & 2.99 & 35.00 & 74.00 & 52.65 & 61.52 & 44.43 & 68.78 & 41.95 & 52.11 & 35.24 \\
    \rowcolor{gray!30}
    BAN-BiT & ResNet-18 & 3.80 & 330.32 & 76.84 & 63.19 & 69.35 & 53.08 {\color{green!50!black}$\uparrow$\scriptsize{8.65}} & 85.29 & 51.38 & 64.13 & 47.20 {\color{green!50!black}$\uparrow$\scriptsize{11.96}} \\
    CF* \cite{bandara2022transformer} & MiT-b0 & 3.85 & 11.38 & 76.73 & 57.33 & 65.63 & 48.84 & 74.26 & 46.79 & 57.41 & 40.26 \\
    \rowcolor{gray!30}
    BAN-CF & MiT-b0 & 4.34 & 300.17 & 77.32 & 64.18 & 70.14 & 54.02 {\color{green!50!black}$\uparrow$\scriptsize{5.18}} & 75.65 & 54.35 & 63.25 & 46.26 {\color{green!50!black}$\uparrow$\scriptsize{6.00}} \\
    \hline
    CF-SCD* \cite{xie2021segformer} & MiT-b0 & 4.24 & 20.69 & 77.93 & 61.35 & 68.65 & 52.27 & 75.77 & 53.71 & 62.86 & 45.84 \\
    \rowcolor{gray!30}
    BAN-CF-SCD & MiT-b0 & 5.40 & 316.26 & 78.19 & 67.71 & 72.57 & 56.95 {\color{green!50!black}$\uparrow$\scriptsize{4.68}} & 75.49 & 59.68 & 66.66 & 50.00 {\color{green!50!black}$\uparrow$\scriptsize{4.16}} \\
    CF-SCD* \cite{xie2021segformer} & MiT-b2 & 25.51 & 55.93 & 79.53 & 67.14 & 72.81 & 57.24 & 76.58 & 56.70 & 65.15 & 48.32 \\
    \rowcolor{gray!30}
    BAN-CF-SCD & MiT-b2 & 28.31 & 354.59 & 79.66 & 70.44 & 74.77 & 59.70 {\color{green!50!black}$\uparrow$\scriptsize{2.46}} & 76.08 & 61.33 & 67.91 & 51.41 {\color{green!50!black}$\uparrow$\scriptsize{3.09}} \\
    \bottomrule[1pt]
  \end{tabular}}
\end{table*}

BCD is the fundamental task of CD, and we conducted fully-supervised BCD experiments on three datasets including LEVIR-CD, S2Looking, and BANDON using only BCD labels (``BANDON-BCD'' for short). As listed in \cref{table_levircd}, we re-implemented BiT and the results obtained are better than those reported in the original paper\footnote[1]{Some results in \cref{table_levircd}, \cref{table_s2looking} and \cref{table_bandon} follow \cite{chen2021remote}, \cite{shen2021s2looking} and \cite{pang2023detecting}.}, but still not better than the latest 6 customized models. After using BiT as the Bi-TAB in our BAN framework, it achieves up to 2.68\% improvement in $IoU^c$, with only 0.25M learnable parameter cost. Similarly, after injecting general knowledge, CF achieves 91.86\% in $F_1^c$ score and 84.94\% in $IoU^c$, achieving the best performance among the compared methods, and our implementation of CF-b0 has far fewer parameters than the original CF. In addition, we presented a simple light-weight CD model, \textit{stacked-blocks}, using several hand-crafted stacked blocks under the parameter number of only 1.44M, which achieves 3.23\% $IoU^c$ improvement under the framework of BAN. This improvement further demonstrates that general knowledge can provide considerable gains under limited conditions.

For the more challenging S2Looking dataset, BiT and CF-b1 also achieve the best composite metrics after equipping to BAN, which brings 1.83\% and 2.36\% improvement in $F_1^c$ and $IoU^c$, as listed in \cref{table_s2looking}. Also for off-nadir aerial images, BANDON-BCD focuses on urban scenes rather than rural scenes in S2Looking. On the BANDON-BCD dataset, as listed in \cref{table_bandon}, the performance of the plain BiT is inferior to that of almost all the compared methods. Driven by BAN, the performance of BiT is dramatically improved, with $IoU^c$ spanning 8.65\% from 44.43\% to 53.08\%. For CF, the $IoU^c$ is improved from 48.84\% to 54.02\%, which achieves the best BCD performance when not using additional auxiliary annotations.

To better analyze where BAN brings improvements, we performed some visualizations on the LEVIR-CD dataset, as shown in \cref{fig:cd_compare}, where we use red and yellow to render missed and false detections. We found that BAN has better detection performance on buildings with a similar appearance to the background, such as the circled areas in \cref{fig:cd_compare}(a)$\sim$(e). In addition, for some changes in buildings with a special appearance, they can be better detected after applying BAN, as shown by the purple buildings in \cref{fig:cd_compare}(f). A common feature of the above scenes is that they have a relatively small number of samples in the dataset, and thus, to some extent, we can believe that the injection of general knowledge enables the model to achieve a stronger ability to learn from fewer samples. Another interesting observation is that in \cref{fig:cd_compare}(g), models without BAN often mis-recognize cargo box changes as building changes due to their similar appearance. With only limited change detection datasets, it is difficult for the model to learn to distinguish between cargo boxes and buildings, but the introduction of general knowledge makes it possible.

\subsubsection{\textbf{Semantic Change Detection}}

\begin{table}
  \caption{Comparisons of BAN with other SCD methods on learnable parameters ($Param$ (M)), $mIoU$ (\%), $Sek$ (\%) and $Score$ (\%) on Landsat-SCD dataset. The symbol ``*'' means our re-implemented results.}
  \label{table_lansat}
  \centering
  \scalebox{0.9}{
  \begin{tabular}{@{}l|c|c|c|c@{}}
    \toprule[1pt]
    Method & $Param$ & $mIoU$ & $Sek$ & $Score$\\
    \midrule
    HRSCDstr3 \cite{daudt2019multitask} & N/A & 78.15 & 31.64 & 45.60 \\
    HRSCDstr4 \cite{daudt2019multitask} & 62.68 & 79.15 & 37.03 & 49.66 \\
    SCDNet \cite{peng2021scdnet} & 30.01 & 70.17 & 39.69 & 51.83 \\
    Bi-SRNet \cite{ding2022bi} & 23.38 & 82.02 & 42.60 & 54.43 \\
    SSCDI \cite{ding2022bi} & 23.31 & 82.10 & 42.75 & 54.56 \\
    \hline
    CF-SCD-b0* \cite{xie2021segformer} & 4.24 & 83.99 & 52.69 & 62.08 \\
    \rowcolor{gray!30}
    BAN-CF-SCD-b0 & 5.40 & 84.76 & 54.56 & 63.62 {\color{green!50!black}$\uparrow$\scriptsize{1.54}} \\ 
    CF-SCD-b1* \cite{xie2021segformer} & 14.70 & 85.99 & 55.69 & 64.78 \\
    \rowcolor{gray!30}
    BAN-CF-SCD-b1 & 17.27 & 86.41 & 58.56 & 66.92 {\color{green!50!black}$\uparrow$\scriptsize{2.14}} \\
    \bottomrule[1pt]
  \end{tabular}}
\end{table}

\begin{table}
  \caption{Comparisons of BAN with other SCD methods on learnable parameters ($Param$ (M)), $mIoU$ (\%), $Sek$ (\%) and $Score$ (\%) on BANDON In-domain testing set. The symbol ``*'' means our re-implemented results.}
  \label{table_bandon_scd}
  \centering
  \scalebox{0.9}{
  \begin{tabular}{@{}l|c|c|c|c@{}}
    \toprule[1pt]
    Method & $Param$ & $mIoU$ & $Sek$ & $Score$\\
    \midrule
    CF-SCD-b0* \cite{xie2021segformer} & 4.24 & 75.27 & 27.95 & 42.15 \\
    \rowcolor{gray!30}
    BAN-CF-SCD-b0 & 5.40 & 77.68 & 31.48 & 45.34 {\color{green!50!black}$\uparrow$\scriptsize{3.19}} \\ 
    CF-SCD-b2* \cite{xie2021segformer} & 25.51 & 77.84 & 33.49 & 46.79 \\
    \rowcolor{gray!30}
    BAN-CF-SCD-b2 & 28.31 & 79.11 & 35.20 & 48.37 {\color{green!50!black}$\uparrow$\scriptsize{1.58}} \\
    \bottomrule[1pt]
  \end{tabular}}
\end{table}

SCD is a derivative task of BCD which not only detects where the change has occurred but also predicts \textbf{from} which class \textbf{to} which class. We conducted SCD experiments on two datasets, including the Landsat dataset and the BANDON dataset using SCD annotation (i.e., additional semantic segmentation labels). As listed in \cref{table_lansat}, in the Landsat data set, CF-SCD-b0 and CF-SCD-b1 are used as Bi-TAB of BAN and achieve a 1.54\% and 2.14\% improvement in the weighted $Score$. It is worth noting that these two models have an improvement of 1.87\% and 2.87\% in the $Sek$ metric, indicating that BAN is not only effective in the CD task, but also drives the semantic segmentation subtask to obtain better performance, and the same observation can be made in experiments on the BANDON-SCD dataset in \cref{table_bandon_scd}. The results of the auxiliary semantic segmentation task of BANDON-SCD are visualized in \cref{fig:cd_compare_bandon}, where we found that BAN has a more accurate detection of ambiguous targets, which may be caused by blurring, noise and entanglement of the image with the background, etc., as shown in \cref{fig:cd_compare_bandon}(a)$\sim$(c).

The metrics in \cref{table_bandon} focus only on CD performance, ignoring the performance of the semantic segmentation auxiliary task in \cref{table_bandon_scd}. On the BANDON-SCD dataset, when only using our reconstructed CF-SCD-b0, 68.65\% $F_1^c$ and 52.27\% $IoU^c$ are obtained, which is inferior to Change-Mix \cite{zheng2021change}. After equipping to the BAN, i.e. BAN-CF-SCD-b0, it achieves 56.95\% $IoU^c$, with 4.68\% improvement. Compared to BAN-CF-B0 without SCD annotation, there is a 2.93\% improvement. With a larger Bi-TAB, i.e., MiT-b2, BAN-CF-SCD can achieve 59.70\% IoU, which is only 0.57\% inferior to MTGCD-Net \cite{pang2023detecting}. Note that MTGCD-Net uses more auxiliary tasks with annotations, including the building roof-to-footprint offsets and the bi-temporal matching flows between identical roofs, so it is reasonable to achieve the best performance.

\subsubsection{\textbf{Cross-domain Change Detection}}

Cross-domain CD is a critical issue in practical applications of CD. In practice, it is expensive to annotate specialized data for each scene and train scene-specific models. Therefore, we generally expect the model to have good performance not only on in-domain datasets but also on out-domain datasets, i.e. strong generalizability. To explore the effectiveness of BAN on out-domain data, we validated it on the out-domain testing set of the BANDON dataset. As listed in \cref{table_bandon}, the plain BiT achieves only 35.24\% $IoU^c$ on the out-domain data, which is 9.19\% lower than that on the in-domain data, which indicates that cross-domain inference drastically affects the customized CD model. After equipping to BAN, BAN-BiT achieves 11.96\% improvement, although the cross-domain damage still exists, it is effectively mitigated. Similarly, BAN-CF achieves a 6.00\% improvement over CF on the out-domain testing set. Compared to the in-domain testing set, the magnitude of the improvement brought by BAN is larger, which suggests that BAN contributes to the generalization ability of the model. We also conducted cross-domain CD experiments on the BANDON-SCD dataset, and after equipping to BAN, BAN-CF-SCD-b0 and b2 obtained similar improvements as on the in-domain testing set, 4.16\% and 3.09\%, respectively.

\subsubsection{\textbf{Semi-supervised Change Detection}}

\begin{table*}
  \caption{Comparisons of BAN with other semi-supervised CD methods on $IoU^c$ (\%) and $OA$ (\%) on WHU-CD dataset.}
  \label{table_whucd}
  \centering
  \scalebox{0.95}{
  \begin{tabular}{@{}l|cc|cc|cc|cc@{}}
    \toprule[1pt]
    \multirow{2}{*}{Method} & \multicolumn{2}{c|}{5\%} & \multicolumn{2}{c|}{10\%} & \multicolumn{2}{c|}{20\%} & \multicolumn{2}{c}{40\%}\\
    & $IoU^c$ & $OA$ & $IoU^c$ & $OA$ & $IoU^c$ & $OA$ & $IoU^c$ & $OA$\\
    \midrule
    Sup. only \cite{bandara2022revisiting} & 50.00 & 97.48 & 55.70 & 97.53 & 65.40 & 98.20 & 76.10 & 98.04 \\
    AdvNet \cite{vu2019advent} & 55.10 & 97.90 & 61.60 & 98.11 & 73.80 & 98.80 & 76.60 & 98.94 \\
    s4GAN \cite{mittal2019semi} & 18.30 & 96.69 & 62.60 & 98.15 & 70.8 & 98.60 & 76.40 & 98.96 \\
    SemiCDNet \cite{peng2020semicdnet} & 51.70 & 97.71 & 62.00 & 98.16 & 66.70 & 98.28 & 75.90 & 98.93 \\
    SemiCD \cite{bandara2022revisiting} & \textbf{65.80} & \textbf{98.37} & \textbf{68.10} & 98.47 & 74.80 & 98.84 & 77.20 & 98.96 \\
    \rowcolor{gray!30}
    \textbf{ours} & 59.21 {\color{red!50!black}$\downarrow$\scriptsize{6.59}} & 98.32 & 64.91 {\color{red!50!black}$\downarrow$\scriptsize{3.19}} & \textbf{98.54} & \textbf{81.05} {\color{green!50!black}$\uparrow$\scriptsize{6.25}} & \textbf{99.21} & \textbf{85.26} {\color{green!50!black}$\uparrow$\scriptsize{8.06}} & \textbf{99.39} \\
    \bottomrule[1pt]
  \end{tabular}}
\end{table*}

In Section \ref{section:bcd_exp}, we found that BAN has a strong ability to learn from a few samples through visualization, and thus we further validated this observation here with some semi-supervised CD settings. Semi-supervised CD refers to training the CD model with a small amount of labeled data and a large amount of unlabeled data. Here, we did not train BAN on unlabeled data, but just followed some semi-supervised settings, i.e., training with partial labels. As listed in \cref{table_whucd}, we compared BAN with several semi-supervised learning methods on the WHU-CD dataset, where AdvNet \cite{vu2019advent} and s4GAN \cite{mittal2019semi} are semi-supervised semantic segmentation methods, SemiCDNet \cite{peng2020semicdnet} and SemiCD \cite{bandara2022revisiting} are state-of-the-art semi-supervised CD methods, and ``Sup. only'' is reported in \cite{bandara2022revisiting}, which only performs fully supervised learning on the labeled data. When using only 5\% and 10\% of the labeled data, BAN is superior to other fully supervised and semi-supervised methods except for SemiCD, which is acceptable because BAN does not use any unlabeled data. When using 20\% of the labeled data, BAN achieves 81.05\% $IoU^c$, which is superior to SemiCD by 6.25\%. When using 40\% of the labeled data, BAN achieves 85.26\% $IoU^c$, 8.06\% superior to SemiCD. These results show us the possibility and potential of BAN for future applications in semi-supervised learning frameworks.

\subsection{Ablation Studies}
\label{section:ablation_studies}

\subsubsection{ARIS}

\begin{table}
  \caption{Performance of BAN at different input resolutions for foundation models (ViT-L/14 of CLIP and RemoteCLIP) on $F_1^c$ (\%), $IoU^c$ (\%) and FPS (img/s) on LEVIR-CD dataset.}
  \label{table_resolution}
  \centering
  \scalebox{0.9}{
  \begin{tabular}{@{}l|cc|cc|c@{}}
    \toprule[1pt]
    \multirow{2}{*}{Resolution} & \multicolumn{2}{c|}{ViT-L/14 (CLIP)} & \multicolumn{2}{c|}{ViT-L/14 (RemoteCLIP)} & \multirow{2}{*}{FPS}\\
    & $F_1^c$ & $IoU^c$ & $F_1^c$ & $IoU^c$ & \\
    \midrule
    $224 \times 224$ & 91.68 & 84.63 & \cellcolor{gray!30}\textbf{91.92} & \cellcolor{gray!30}\textbf{85.05} & 2.86 \\
    \textbf{$336 \times 336$} & \cellcolor{gray!30}\textbf{91.86} & \cellcolor{gray!30}\textbf{84.94} & 91.80 & 84.85 & 1.79 \\
    $512 \times 512$ & 91.77 & 84.79  & 91.84 & 84.92 & 0.71\\
    \bottomrule[1pt]
  \end{tabular}}
\end{table}

To avoid recomputation of the position encoding in the foundation model and the resolution gap during training and inferring, ARIS is adopted. We conducted this ablation on the LEVIR-CD dataset using CLIP's ViT-L/14 (pretrained on $336 \times 336$ images) and RemoteCLIP's ViT-L/14 (pretrained on $224 \times 224$ images) as the foundation model. And the input resolution of the Bi-TAB is kept at $512 \times 512$. As listed in \cref{table_resolution}, when using the low input resolution, i.e., $224 \times 224$, the model achieves the best Frames Per Second (FPS), with 2.86 images of $1024 \times 1024$ per second inferred using the sliding-window when the batch size is 1 on a single A6000 GPU. When the input resolution matched that of the pre-training, i.e. $224 \times 224$ for RemoteCLIP's ViT-L/14 or $336 \times 336$ for CLIP's ViT-L/14, is applied, the model achieves the best performance. When keeping the constant input resolution, i.e., $512 \times 512$, the FPS is only 0.71 and its performance is inferior to that when using ARIS.

\subsubsection{Different Foundation Models}
\label{section:ablation_studies_dfm}

\begin{table}
  \caption{Performance of BAN with different foundation models on $Precision^c$($P^c$) (\%), $Recall^c$($R^c$) (\%), $F_1^c$ (\%) and $IoU^c$ (\%) on LEVIR-CD dataset.}
  \label{table_fm}
  \centering
  \scalebox{0.9}{
  \begin{tabular}{@{}l|c|c|c|c|c@{}}
    \toprule[1pt]
    Foundation Model & Pre-train & $P^c$ & $R^c$ & $F_1^c$ & $IoU^c$\\
    \midrule
    ViT-B/16 & IN-21k, sup \cite{steiner2021train} & 93.59 & 89.80 & 91.66 & 84.60 \\
    ViT-L/16 & IN-21k, sup \cite{steiner2021train} & 93.27 & 90.11 & 91.67 & 84.61 \\
    \hline
    ViT-B/16 & CLIP \cite{cherti2023reproducible} & 93.25 & 90.21 & 91.71 & 84.68 \\
    ViT-L/14 & CLIP \cite{cherti2023reproducible} & 93.47 & 90.30 & 91.86 & 84.94 \\
    \hline
    ViT-B/32 & RemoteCLIP \cite{liu2023remoteclip} & 93.28 & 90.26 & 91.75 & 84.75 \\
    ViT-L/14 & RemoteCLIP \cite{liu2023remoteclip} & 93.44 & 90.46 & 91.92 & 85.05 \\
    \hline
    ViT-B/32 & GeoRSCLIP \cite{zhang2023rs5m} & 93.35 & 90.24 & 91.77 & 84.79 \\
    ViT-L/14 & GeoRSCLIP \cite{zhang2023rs5m} & 93.50 & 90.48 & 91.96 & 85.13 \\
    \hline
    InternImage-XL \cite{wang2023internimage} & IN-21k, sup & 93.53 & 90.41 & 91.94 & 85.09 \\
    \bottomrule[1pt]
  \end{tabular}}
\end{table}

With differences in training data and training strategies, various foundation models contain different types of general knowledge. Therefore, we used different foundation models to explore the impact of different general knowledge sources on the performance of BAN. Specifically, for the ViT-based foundation models, we used ViTs fully supervised trained on ImageNet-21k, ViTs trained on the LAION-2B dataset using contrastive language-image learning (CLIP), and ViTs trained on RS data (RemoteCLIP, GeoRSCLIP), respectively, as the general knowledge sources in the BAN. As listed in \cref{table_fm}, for the foundation models using the same pre-training, the large-scale models generally perform better than the small-scale models. For the same scale, the model trained using CLIP is superior to the model fully supervised trained on ImageNet-21k. In addition, the CLIP model pre-trained on RS data is superior to models of the corresponding scale trained on non-RS data, which is in part due to the fact that general knowledge within the RS domain is easier to transfer and adapt to the CD task of RS images. We also attempted to use the deformable convolution-based model (i.e. InternImage), which is trained in a supervised manner on ImageNet-21k, as the foundation model in BAN, and the results demonstrate that the structure of the foundation model in BAN is not limited to ViT and that BAN can benefit from stronger foundation models.

\subsubsection{Scalable Bi-TAB}

\begin{table}
  \caption{Performance of BAN with scalable Bi-TABs on learnable parameters ($Param$ (M)), $Precision^c$($P^c$) (\%), $Recall^c$($R^c$) (\%), $F_1^c$ (\%) and $IoU^c$ (\%) on LEVIR-CD dataset.}
  \label{table_bitab}
  \centering
  \scalebox{0.9}{
  \begin{tabular}{@{}l|c|c|c|c|c@{}}
    \toprule[1pt]
    Method & $Param$ & $P^c$ & $R^c$ & $F_1^c$ & $IoU^c$\\
    \midrule
    CF \cite{bandara2022transformer} & 41.02 & 92.05 & 88.80 & 90.40 & 82.48 \\
    BAN-CF-b0 & 4.34 & 93.47 & 90.30 & 91.86 & 84.94 {\color{green!50!black}$\uparrow$\scriptsize{2.46}} \\
    BAN-CF-b1 & 15.88 & 93.48 & 90.76 & 92.10 & 85.36 {\color{green!50!black}$\uparrow$\scriptsize{2.88}} \\
    BAN-CF-b2 & 26.92 & 93.61 & 91.02 & 92.30 & 85.69 {\color{green!50!black}$\uparrow$\scriptsize{3.21}} \\
    \bottomrule[1pt]
  \end{tabular}}
\end{table}

In \cref{section:Bi-TAB}, it is mentioned that compared to some stacked blocks, we recommend using the existing customized CD models as Bi-TAB, which allows BAN to benefit from these evolving models. In \cref{table_bitab}, we used scalable CFs as Bi-TAB and found that BAN can significantly improve the performance after using more complex Bi-TABs. Compared to CF, BAN-CF-b0 achieves an improvement of 2.46\% $IoU^c$, BAN-CF-b1 achieves an improvement of 2.88\% $IoU^c$, and BAN-CF-b2 achieves an improvement of 3.21\% $IoU^c$. This trend further suggests that it is reasonable to consider Bi-TAB as a model-agnostic concept.

\section{Conclusion}

In this paper, we propose a universal framework (i.e. BAN) to utilize general knowledge from large foundation models to reduce the dependence of existing CD models on a large amount of labeled data. Experimental results demonstrate that with only a few additional learnable parameters, BAN can effectively improve the performance of existing CD methods and enable them to learn better from fewer samples. Since both the foundation model and Bi-TAB in BAN are model-agnostic, BAN is highly extensible and can benefit from stronger foundation models, thus helping boost the existing customized CD models. More importantly, our BAN confirms the feasibility of foundation model adaptation in CD tasks and provides the research basis for subsequent investigations. In the future, we suggest further exploration of BAN or foundation model-based CD methods on several aspects, including, but not limited to, better PETL methods, more effective bridging of general and task-specific features, and more suitable task-specific modules. In addition, as a universal framework, BAN is not limited to CD on RGB images, and it can be easily extended to spectral data, i.e., multispectral and hyperspectral images.


\bibliographystyle{IEEEtran}
\bibliography{IEEEabrv,BAN.bib}

\vfill

\end{document}